\documentclass[a4paper]{article}

\usepackage{INTERSPEECH2018}
\usepackage{CJKutf8}
\usepackage[utf8]{inputenc}
\usepackage[T1]{fontenc}
\usepackage{booktabs}
\usepackage{multirow}
\usepackage{mathtools}
\usepackage{amsmath,graphicx,caption}
\usepackage{xcolor}
\usepackage{xspace}
\usepackage{paralist}
\usepackage{url}
\usepackage{enumitem}
\usepackage{multicol}
\usepackage[colorinlistoftodos]{todonotes}
\usepackage{subcaption}
\usepackage[export]{adjustbox}
\setlist[itemize]{leftmargin=*}
\usepackage{footnote}
\makesavenoteenv{tabular}
\makesavenoteenv{table}

\usepackage[symbol]{footmisc}
\renewcommand{\thefootnote}{\fnsymbol{footnote}}

\renewcommand{\paragraph}[1]{\medskip\noindent\textbf{#1}~}

\newcommand{\sos}{\ensuremath{\mathtt{\langle s\rangle}}\xspace}
\newcommand{\eos}{\ensuremath{\mathtt{\langle /s\rangle}}\xspace}
\newcommand{\sw}{\ensuremath{\mathtt{\langle sw\rangle}}\xspace}
\newcommand{\vocab}[1]{\ensuremath{\mathrm{V_{#1}}}\xspace}
\newcommand{\LM}[1]{\ensuremath{\mathcal{L}_{#1}}\xspace}

\makeatletter
\newcommand\Label[1]{&\refstepcounter{equation}(\theequation)\ltx@label{#1}&}
\makeatother

\newsavebox{\fmbox}
\newenvironment{fmpage}[1]
{\begin{lrbox}{\fmbox}\begin{minipage}{#1}}
{\end{minipage}\end{lrbox}\fbox{\usebox{\fmbox}}}

\title{Dual Language Models for \\Code Switched Speech Recognition}
\name{Saurabh Garg, Tanmay Parekh, Preethi Jyothi}
\address{Department of Computer Science and Engineering, Indian Institute of Technology Bombay}
\email{\{saurabhgarg,tanmayb,pjyothi\}@cse.iitb.ac.in}

\begin{document}

\maketitle
\begin{abstract}
In this work, we present a simple and elegant approach to language modeling for bilingual code-switched text. Since code-switching is a blend of two or more different languages, a standard bilingual language model can be improved upon by using structures of the monolingual language models. We propose a novel technique called {\em dual language models}, which involves building two complementary monolingual language models and combining them using a probabilistic model for switching between the two. We evaluate the efficacy of our approach using a conversational Mandarin-English speech corpus. We prove the robustness of our model by showing significant improvements in perplexity measures over the standard bilingual language model without the use of any external information. Similar consistent improvements are also reflected in automatic speech recognition error rates.
\end{abstract}
\vspace{5pt}
\section{Introduction}

\label{sec:intro}

Code-switching is a commonly occurring phenomenon in multilingual communities, wherein a speaker switches between languages within the span of a single utterance. Code-switched speech presents many challenges for automatic speech recognition (ASR) systems, in the context of both acoustic models and language models. Mixing of dissimilar languages leads to loss of structure, which makes the task of language modeling more difficult.  Our focus in this paper is on building robust language models for code-switched speech from bilingual speakers.

A na\"ive approach towards this problem would be to simply use a bilingual language model. However, the complexity of a full-fledged bilingual language model is significantly higher than that of two monolingual models, and is unsuitable in a limited data setting. More sophisticated approaches relying on translation models have been proposed to overcome this challenge (see Section~\ref{sec:relwork}), but they rely on external resources to build the translation model. In this paper, we introduce an alternate -- and simpler -- approach to address the challenge of limited data in the context of code-switched text without use of any external resources.

At the heart of our solution is a {\em dual language model} (DLM) that has roughly the complexity of two monolingual language models combined. A DLM combines two such models and uses a probabilistic model to switch between them. Its simplicity makes it amenable for generalization in a low-data context. Further there are several other benefits of using DLMs. (1) The DLM construction does not rely on any prior information about the underlying languages. (2) Since the structure of our combined model is derived from monolingual language models, it can be implemented as a finite-state machine and easily incorporated within an ASR system. (3) The monolingual language model for the primary language can be trained further with large amounts of monolingual text data (which is easier to obtain compared to code-switched text).

Our main contributions can be summarized as follows:
\begin{itemize}[topsep=1pt,itemsep=2pt,parsep=0pt,partopsep=0pt]
\item We formalize the framework of DLMs (Section~\ref{sec:dlms}).
\item We show significant improvements in perplexity using DLMs when compared against smoothed $n$-gram language models estimated on code-switched text (Section~\ref{sec:perplexity}). We provide a detailed analysis of DLM improvements (Section~\ref{sec:discussion}).
\item We evaluate DLMs on the ASR task. DLMs capture sufficient complementary information which we leverage to show improvements on error rates. (Section~\ref{sec:ASR}).
\end{itemize}

\begin{figure*}[h!]
\begin{fmpage}{0.98\linewidth}
Given two language models \LM1 and \LM2 with conditional probabilities $P_1$ and $P_2$ that satisfy the following conditions:
\begin{align*}
P_1[\eos\mid\sos] = P_2[\eos\mid\sos] &= 0 \Label{eq:cond1} & 
P_1[\sw\mid\sos] + P_2[\sw\mid\sos] &= 1 \Label{eq:cond2}\\
P_1[\sw\mid\sw] = P_2[\sw\mid\sw] &= 0 \Label{eq:cond3} &
P_1[\eos\mid\sw] = P_2[\eos\mid\sw] &= 0 \Label{eq:cond4}
\end{align*}
We define a combined language model \LM, with conditional probabilities $P$, as follows:
\begin{align*}
P[w'\mid w] &= \begin{cases}
P_1[w'\mid\sos] &\text{ if } w' \in \vocab1 \\
P_2[w'\mid\sos] &\text{ if } w' \in \vocab2 \\
0 &\text{ if } w' = \eos
\end{cases}
& \text{for } w=\sos \\
P[w'\mid w] &= \begin{cases}
P_1[w'\mid w] &\text{ if } w' \in \vocab1 \cup \{\eos\} \\
P_1[\sw\mid w]\cdot P_2[w'\mid \sw] &\text{ if } w' \in \vocab2
\end{cases}
& \text{for } w \in \vocab1 \\
P[w'\mid w] &= \begin{cases}
P_2[w'\mid w] &\text{ if } w' \in \vocab2 \cup \{\eos\} \\
P_2[\sw\mid w]\cdot P_1[w'\mid \sw] &\text{ if } w' \in \vocab1
\end{cases}
& \text{for } w \in \vocab2
\end{align*}
\end{fmpage}
\caption{Definition of a bigram-based DLM for code-switched text.}
\label{fig:algo}
\vspace{-1em}
\end{figure*}

\section{Related Work}
\label{sec:relwork}

Prior work on building ASR systems for code-switched speech can be broadly categorized into two sets of approaches: (1) Detecting code-switching points in an utterance, followed by the application of monolingual acoustic and language models to the individual segments~\cite{chan2004detection,lyu2008language,shia2004language}. (2) Employing a universal phone set to build acoustic models for the mixed speech and pairing it with standard language models trained on code-switched text~\cite{imseng2011language,li2011asymmetric,bhuvanagiri2010approach,yeh2010integrated,yu2003chinese}. 

There have been many past efforts towards enhancing the capability of language models for code-switched speech using additional sources of information such as part-of-speech (POS) taggers and statistical machine translation (SMT) systems. Yeh et al.~\cite{yeh2010integrated} employed class-based $n$-gram models that cluster words from both languages into classes based on POS and perplexity-based features. Vu et al.~\cite{vu2012first} used an SMT system to enhance the language models during decoding. Li et al.~\cite{li2013improved} propose combining a code-switch boundary predictor with both a translation model and a reconstruction model to build language models. (Solorio et. al.~\cite{solorio2008learning} were one of the first works on learning to predict code-switching points.) Adel et al.~\cite{adel2015syntactic} investigated how to effectively use syntactic and semantic features extracted from code-switched data within factored language models. Combining recurrent neural network-based language models with such factored language models has also been explored~\cite{adel2014combining}.

\section{Dual language models}
\label{sec:dlms}

We define a {\em dual language model} (DLM) to have the following 2-player game structure. A sentence  (or more generally, a sequence of tokens) is generated via a co-operative game between the two players who take turns. During its turn a player generates one or more words (or tokens), and either terminates the sentence or transfers control to the other player. Optionally, while transferring control, a player may send additional information to the other player (e.g., the last word it produced), and also may retain some state information (e.g., cached words) for its next turn. At the beginning of the game one of the two players is chosen probabilistically.

In the context of code-switched text involving two languages, we consider a DLM wherein the two players are each in charge of generating tokens in one of the two languages. Suppose the two languages have (typically disjoint) vocabularies \vocab1 and \vocab2. Then the alphabet of the output tokens produced by the first player in a single turn is $\vocab1 \cup \{ \sw, \eos \}$, \sw denotes the {\em switching} -- i.e., transferring control to the other player -- and \eos denotes the end of sentence, terminating the game. We shall require that a player produces at least one token before switching or terminating, so that when $\vocab1\cap\vocab2=\emptyset$, any non-empty sentence in $(\vocab1 \cup \vocab2)^*$ uniquely determines the sequence of corresponding outputs from the two players when the DLM produces that sentence. (Without this restriction, the players can switch control between each other arbitrarily many times, or have either player terminate a given sentence.)

In this paper, we explore a particularly simple DLM that is constructed from two given LMs for the two languages. More precisely, we shall consider an LM \LM1 which produces \eos-terminated strings in $(\vocab1 \cup \{ \sw \})^*$ where \sw indicates a span of tokens in the {\em other} language (so multiple \sw tokens cannot appear adjacent to each other), and symmetrically an LM \LM2 which produces strings in  $(\vocab2 \cup \{ \sw \})^*$. 
In Section~\ref{sec:monolm}, we will describe how such monolingual LMs can be constructed from code-switched data. Given \LM1 and \LM2, we shall splice them together into a simple DLM (in which players do not retain any state between turns, or transmit state information to the other player at the end of a turn). Below we explain this process which is formally described in Fig.~\ref{fig:algo} (for bi-gram language models).

We impose conditions \eqref{eq:cond1}-\eqref{eq:cond4} on the given LMs. Condition \eqref{eq:cond1} which disallows empty sentences in the given LMs (and the resulting LM) is natural, and merely for convenience. Condition \eqref{eq:cond2} states the requirement that \LM1 and \LM2 agree on the probabilities with which each of them gets the first turn. Conditions \eqref{eq:cond3} and \eqref{eq:cond4} require that after switching at least one token should be output before switching again or terminating. If the two LMs are trained on the same data as described in Section~\ref{sec:monolm}, all these conditions would hold. 

To see that $P[w'\mid w]$ defined in Fig.~\ref{fig:algo} is a well-defined probability distribution, we check that $\sum_{w'} P[w'|w] = 1$ for all three cases of $w$, where the summation is over $w'\in \vocab1 \cup \vocab2  \cup \{\eos\}$. When $w=\sos$, $\sum_{w'} P[w'|w]$  equals
\begin{align*} 
&\sum_{w'\in \vocab1}P_1[w'|\sos] + \sum_{w'\in \vocab2}P_2[w'|\sos] \\
&\qquad= (1-P_1[\sw\mid\sos]) + (1-P_2[\sw\mid\sos]) = 1
\end{align*}
where the first equality is from~\eqref{eq:cond1} and the second equality is from~\eqref{eq:cond2}. 

When $w \in \vocab1$, $\sum_{w'} P[w'|w]$ is
\begin{align*}
&\sum_{w'\in \vocab1 \cup \eos}P_1[w'|w] + P_1[\sw\mid w]\sum_{w'\in \vocab2}P_2[w'|\sw] \\
&\qquad = \sum_{w'\in \vocab1 \cup \eos}P_1[w'|w] + P_1[\sw\mid w] = 1.
\end{align*}
The case of $w \in \vocab2$ follows symmetrically.

\begin{figure}[t!]
\centering
\includegraphics[scale=.5,width=\columnwidth]{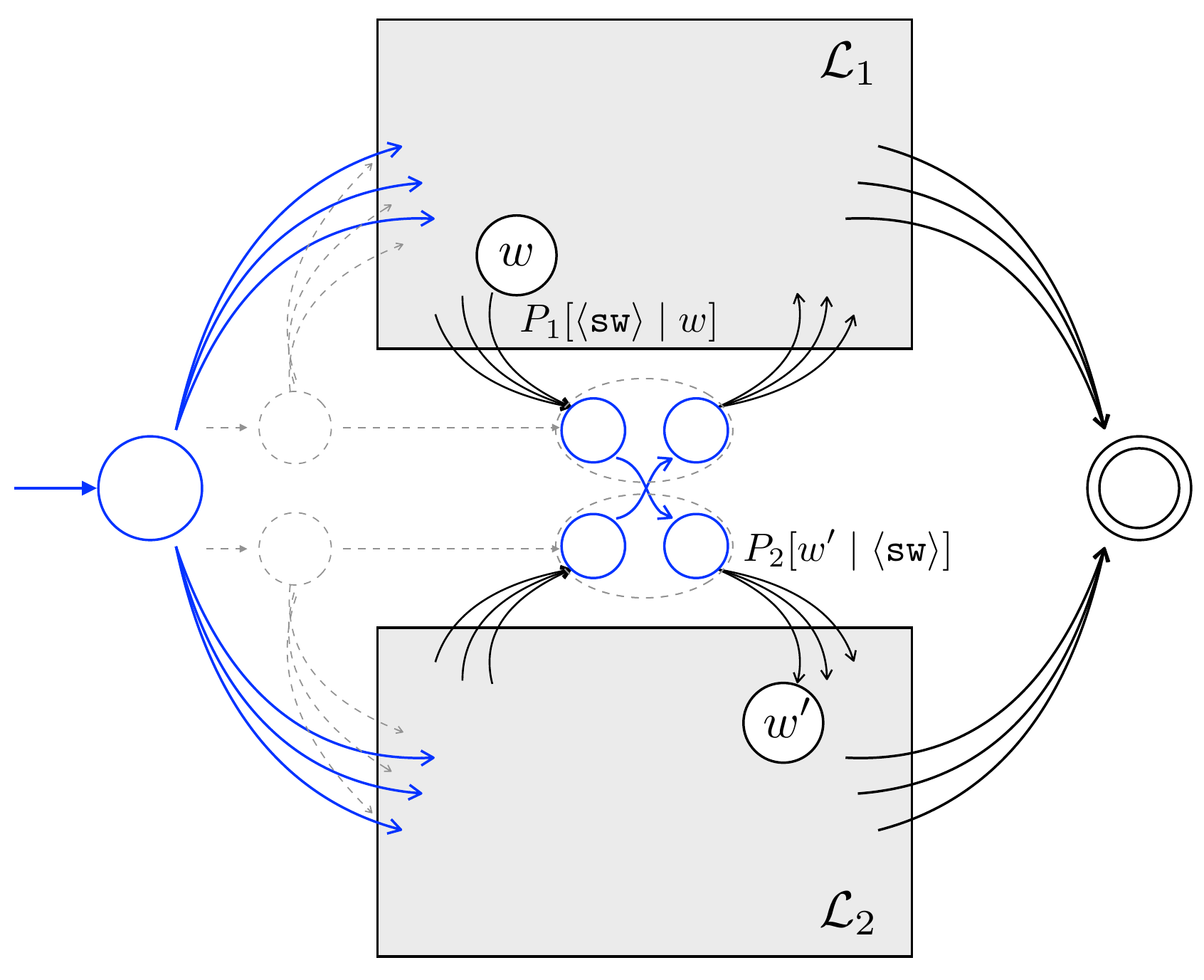}
\caption{DLM using two monolingual LMs, $\LM1$ and $\LM2$, implemented as a finite-state machine.}
\label{fig:fst}
\end{figure}

Figure~\ref{fig:fst} illustrates how to implement a DLM as a finite-state machine using finite-state machines for the monolingual bigram LMs, $\LM1$ and $\LM2$. The start states in both LMs, along with all the arcs leaving these states, are deleted; a new start state and end state is created for the DLM with accompanying arcs as shown in Figure~\ref{fig:fst}. The two states maintaining information about the $\sw$ token can be split and connected, as shown in Figure~\ref{fig:fst}, to create paths between \LM1 and \LM2.
\vspace{-5pt}
\section{Experiments and Results}
\subsection{Data description}
We make use of the SEAME corpus~\cite{lyu2010analysis} which is a conversational Mandarin-English code-switching speech corpus. 

\paragraph{Preprocessing of data.}
\label{sec:preprocessing}
Apart from the code-switched speech, the SEAME corpus comprises of a) words of foreign origin (other than Mandarin and English) b) incomplete words c) unknown words labeled as $\langle unk \rangle$, and d) mixed words such as \begin{CJK*}{UTF8}{gbsn} bleach跟, cause就是, etc.\end{CJK*}. Since it was difficult to obtain pronunciations for these words, we removed utterances that contained any of these words. A few utterances contained markers for non-speech sounds like laughing, breathing, etc. Since our focus in this work is to investigate language models for code-switching, ideally without the interference of these non-speech sounds, we excluded these utterances from our task. 

\paragraph{Data distribution.}
\label{sec:datasplit}
We construct training, development and test sets from the preprocessed SEAME corpus data using a 60-20-20 split. Table~\ref{tab:data} shows detailed statistics of each split. The development and evaluation sets were chosen to have 37 and 30 random speakers each, disjoint from the speakers in the training data.%
\footnote{We note that choosing fewer speakers in the development and test sets led to high variance in the observed results.}
%
The out-of-vocabulary (OOV) rates on the development and test sets are 3.3\% and 3.7\%, respectively.


\begin{table}[h!]
\centering
 \begin{tabular}[b]{ | c | c | c | c | }
 \hline
 {} & Train & Dev & Test \\
 \hline\hline
 \# Speakers & 90 & 37 & 30 \\ 
 \hline
 Duration (hrs) & 56.6 & 18.5 & 18.7 \\
 \hline
 \# Utterances & 54,020 & 19,976 & 19,784 \\
 \hline
 \# Tokens & 539,185 & 195,551 & 196,462 \\
 \hline
\end{tabular}
\captionof{table}{Statistics of the dataset\label{tab:data}}
\vspace{-20pt}
\end{table}
\subsection{Monolingual LMs for the DLM construction}
\label{sec:monolm}
Given a code-switched text corpus $D$, we will derive two complementary corpora, $D_1$ and $D_2$, from which we construct bigram models \LM1 and \LM2 as required by the DLM construction in Figure~\ref{fig:algo}, respectively. In $D_1$, spans of tokens in the second language are replaced by a single token \sw. $D_2$ is constructed symmetrically. Standard bigram model construction on $D_1$ and $D_2$ ensures conditions~\eqref{eq:cond1} and~\eqref{eq:cond2} in Figure~\ref{fig:algo}. The remaining two conditions may not naturally hold: Even though the data in $D_1$ and $D_2$ will not have consecutive \sw tokens, smoothing operations may assign a non-zero probability for this; also, both LMs may assign non-zero probability for a sentence to end right after a \sw token, corresponding to the sentence having ended with a non-empty span of tokens in the other language. These two conditions are therefore enforced by reweighting the LMs.

\subsection{Perplexity experiments}
\label{sec:perplexity}
We used the SRILM toolkit~\cite{srilm} to build all our LMs. The baseline LM is a smoothed bigram LM estimated using the code-switched text which will henceforth be referred to as {\em mixed LM}. Our DLM was built using two monolingual bigram LMs. (The choice of bigram LMs instead of trigram LMs will be  justified later in Section~\ref{sec:discussion}). 
Table~\ref{tab:PPL} shows the perplexities on the validation and test sets using both Good Turing~\cite{goodturing} and Kneser-Ney~\cite{kneserney} smoothing techniques. DLMs clearly outperform mixed LMs on both the datasets. All subsequent experiments use Kneser-Ney smoothed bigram LMs as they perform better than the Good Turing smoothed bigram LMs.

\begin{table}[h!]
\resizebox{\columnwidth}{!}{%
 \begin{tabular}{ | c | c | c | c | c | }
 \hline
 \multirow{2}{*}{\begin{tabular}{c}Smoothing\\ Technique\end{tabular}} & \multicolumn{2}{c|}{Dev} & \multicolumn{2}{c|}{Test} \\\cline{2-5}
 {} &  mixed LM & DLM & mixed LM & DLM \\
  \hline \hline
 Good Turing & 338.2978 &  \textbf{329.1822} & 384.5164 & \textbf{371.1112} \\
 \hline
 Kneser-Ney & 329.6725 & \textbf{324.9268} & 376.0968 & \textbf{369.9355} \\ 
 \hline
\end{tabular}
}
 \caption{Perplexities on the dev/test sets using mixed LMs and DLMs with different smoothing techniques.} \label{tab:PPL}
 \vspace{-1em}
\end{table}
We also evaluate perplexities by reducing the amount of training data to $\frac12$ or $\frac13$ of the original training data (shown in Table~\ref{tab:PPLonVariedTrain}). As we reduce the training data, the improvements in perplexity of DLM over mixed LM further increase, which  validates our hypothesis that DLMs are capable of generalizing better. Section 5 elaborates this point further.

\begin{table}[h!]
\resizebox{\columnwidth}{!}{%
 \begin{tabular}{ | c | c | c | c | c | }
 \hline
 \multirow{2}{*}{\begin{tabular}{c}Training\\ data\end{tabular}} & \multicolumn{2}{c|}{Dev} & \multicolumn{2}{c|}{Test} \\\cline{2-5}
 {} &  mixed LM & DLM & mixed LM & DLM \\
  \hline \hline
Full & 329.6725 & \textbf{324.9268} & 376.0968 & \textbf{369.9355} \\
 \hline
$1/2$ & 362.0966 & \textbf{350.5860} & 400.5831 & \textbf{389.7618} \\ 
 \hline
$1/3$ & 368.6205 & \textbf{356.012} & 408.562 & \textbf{394.2131} \\ 
 \hline
\end{tabular}
}
\captionof{table}{Kneser-Ney smoothed bigram dev/test set perplexities using varying amounts of training data} \label{tab:PPLonVariedTrain}
\vspace{-15pt}
\end{table}
%






\subsection{ASR experiments}
\label{sec:ASR}


All the ASR systems were built using the Kaldi toolkit~\cite{kaldi}. We used standard mel-frequency cepstral coefficient (MFCC)+delta+double-delta features with feature space maximum likelihood linear regression (fMLLR)~\cite{fMLLR} transforms to build speaker-adapted triphone models with $4200$ tied-state triphones, henceforth referred to as ``SAT'' models. We also build time delay neural network (TDNN ~\cite{TDNN})-based acoustic models using i-vector based features (referred to as ``TDNN+SAT"). %
%
%
Finally, we also re-scored lattices generated by the ``TDNN+SAT" model with an RNNLM \cite{dan2018rescoring} (referred to as ``RNNLM Rescoring"), trained using Tensorflow~\cite{abadi2016tensorflow} on the SEAME training data.
\footnote{This rescoring was implemented using the \texttt{tfrnnlm} binary provided by Kaldi~\cite{kaldi} developers.}
We trained a single-layer RNN with 200 hidden units in the LSTM ~\cite{hochreiter1997long} cell.  

The pronunciation lexicon was constructed from CMUdict~\cite{weide1998cmu} and THCHS30 dictionary~\cite{THCHS30_2015} for English and Mandarin pronunciations, respectively. Mandarin words that did not appear in THCHS30 were mapped into Pinyin using a freely available Chinese to Pinyin converter.%
\footnote{\url{https://www.chineseconverter.com/en/convert/chinese-to-pinyin}}
We manually merged the phone sets of Mandarin and English (by mapping all the phones to IPA) resulting in a phone inventory of size $105$.

To evaluate the ASR systems, we treat English words and Mandarin characters as separate tokens and compute token error rates (TERs) as discussed in \cite{vu2012first}. Table~\ref{tab:TER} shows TERs on the dev/test sets using both mixed LMs and DLMs. DLM performs better or on par with mixed LM and at the same time, captures a significant amount of complementary information which we leverage by combining lattices from both systems. The improvements in TER after combining the lattices are statistically significant (at $p<0.001$) for all three systems, which justifies our claim of capturing complementary information. Trigram mixed LM performance was worse than bigram mixed LM; hence we adopted the latter in all our models (further discussed in Section~\ref{sec:discussion}). This demonstrates that obtaining significant performance improvements via LMs on this task is very challenging.
Table~\ref{tab:TERonHalfTrain}  shows all the TER numbers by utilizing only $\frac12$ of the total training data. The combined models continue to give significant improvements over the individual models. Moreover, DLMs consistently show improvements on TERs compared to mixed LMs in the $\frac12$ training data setting.  

\begin{table}[t!]
\resizebox{\columnwidth}{!}{%
\begin{tabular}{ | c | c | c | c | c |}
 \hline
 ASR system & Data & mixed LM & DLM & combined \\ \hline \hline
 \multirow{2}{*}{SAT} & Dev & 45.59 & 45.59 & \textbf{44.93$^*$} \\ \cline{2-5}
 & Test & 47.43 & 47.48 & \textbf{46.96$^*$} \\ \hline
 \multirow{2}{*}{TDNN+SAT} & Dev & 35.20 & 35.26 & \textbf{34.91$^*$} \\ \cline{2-5}
 & Test & 37.42 & 37.35 & \textbf{37.17$^*$} \\ \hline 
  \multirow{2}{*}{RNNLM Rescoring} & Dev & 34.21 & 34.11 & \textbf{33.85$^*$} \\ \cline{2-5}
 & Test & 36.64 & 36.52 & \textbf{36.37$^*$} \\
 \hline
\end{tabular}
}
\caption{TERs using mixed LMs and DLMs} \label{tab:TER}
\vspace*{-15pt}
\end{table}
\begin{table}[tbh]
\resizebox{\columnwidth}{!}{%
\begin{tabular}{ | c | c | c | c | c |}
 \hline
 ASR system & Data & mixed LM & DLM & combined \\ \hline \hline
 \multirow{2}{*}{SAT} & Dev & 48.48 & 48.17 & \textbf{47.67\footnote[1]{statistically significant improvement (at $p<0.001$)}} \\ \cline{2-5}
 & Test & 49.07 & 49.04 & \textbf{48.52$^*$} \\ \hline
 \multirow{2}{*}{TDNN+SAT} & Dev & 40.59 & 40.48 & \textbf{40.12$^*$} \\ \cline{2-5}
 & Test & 41.34 & 41.32 & \textbf{41.13$^*$} \\ \hline 
  \multirow{2}{*}{RNNLM Rescoring} & Dev & 40.20 & 40.09 & \textbf{39.84$^*$} \\ \cline{2-5}
 & Test & 40.98 & 40.90 & \textbf{40.72$^*$} \\
 \hline
\end{tabular}

}
\caption{TERs with $\frac12$ training data} \label{tab:TERonHalfTrain}
\vspace{-1em}
\end{table}

\vspace{-7pt}
\section{Discussion} \label{sec:discussion}

Code-switched data corpora tend to exhibit very different linguistic characteristics compared to standard monolingual corpora, possibly because of the informal contexts in which code-switched data often occurs, and also possibly because of the difficulty in collecting such data. It is possible that the gains made by our language model are in part due to such characteristics of the corpus we use, SEAME. (We note that this corpus is by far the most predominant one used to benchmark speech recognition techniques for code-switched speech.)

In this section we analyze the SEAME corpus and try to further understand our results in light of its characteristics.

\paragraph{Code-switching boundaries.}
Code-switched bigrams with counts of $\leq 10$ occupy $87.5\%$ of the total number of code-switched bigrams in the training data. Of these, $55\%$ of the bigrams have a count of 1. This suggests that context across code-switching boundaries cannot significantly help a language model built from this data. Indeed, the DLM construction in this work discards such context, in favor of a simpler model.

\begin{figure}[t] 
  \centering
  \begin{subfigure}[b]{0.48\linewidth}
           \includegraphics[width=0.9\linewidth,left]{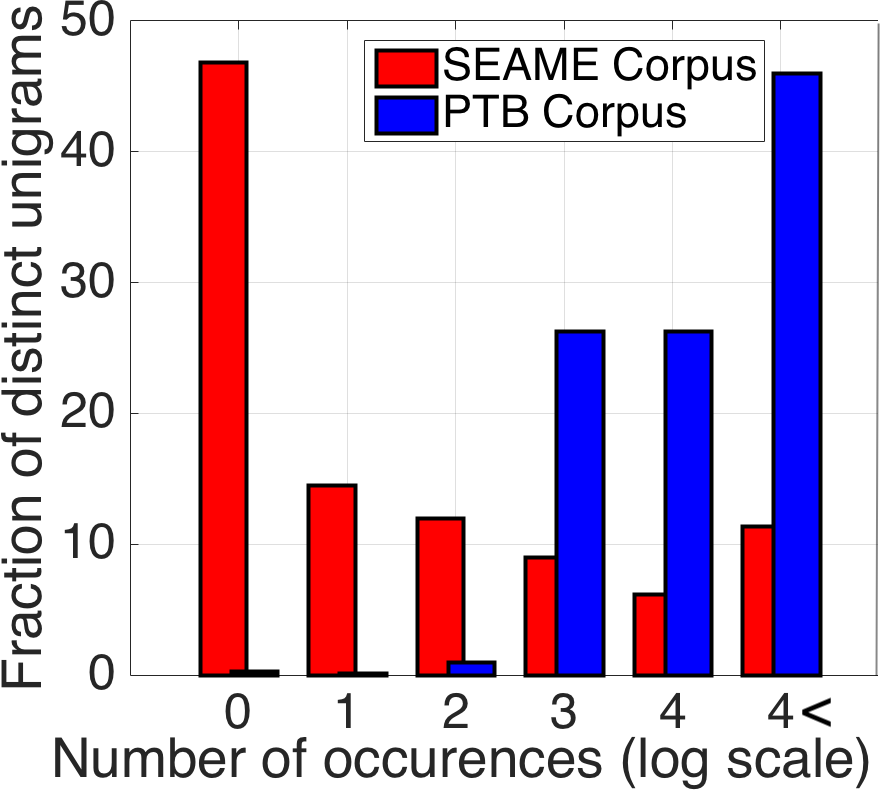}
           \vspace{-15pt}
    \caption{}
    \end{subfigure}
  \begin{subfigure}[b]{0.48\linewidth}
        \includegraphics[width=0.9\linewidth,right]{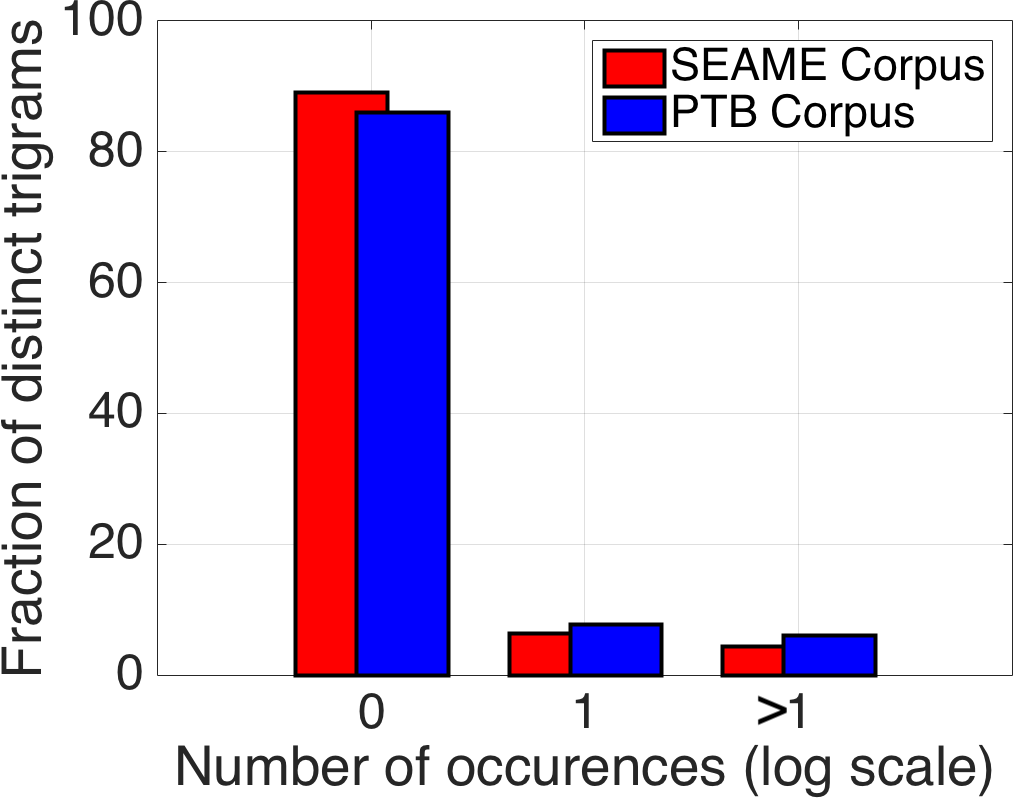}
        \vspace{-15pt}
        \caption{}
    \end{subfigure}
  \caption{Comparison of fraction of data vs. frequency of n-grams in code-mixed text (SEAME) and monolingual English (PTB) text. (The X-axis uses the log2 scale; 1 indicates unigrams with frequencies $\le 2^1$, etc.)}
  \label{fig:ngram}
  \vspace*{-15pt}
\end{figure}

\paragraph{$n$-gram token distribution.} 
We compare the unigram distribution of a code-switched corpus (SEAME) with a standard monolingual corpus (PTB~\cite{marcus1993building}). A glaring difference is observed in their distributions (Figure~\ref{fig:ngram}-a) with significantly high occurrence of less-frequent unigrams in the code-switched corpus, which makes them rather difficult to capture using standard $n$-gram models (which often fall back to a unigram model). The DLM partially compensates for this by emulating a ``class-based language model,'' using the only class information readily available in the data (namely, the language of each word).


\renewcommand*{\thefootnote}{\fnsymbol{footnote}}

\paragraph{Illustrative examples.}
Below, we analyze perplexities of the mixed LM and the DLM on some representative sentences from the SEAME corpus, to illustrate how the performances of the two models compare. 

\begin{CJK*}{UTF8}{gbsn}
\noindent\scalebox{0.93}{\begin{tabular}{p{0.58\linewidth}|c|c}
\toprule
\small
Sentence & \textit{Mixed LM} & \textit{DLM} \\
&  perplexity & perplexity\\
\midrule
我们\ 的\ total\ 是\ 五十七 &  $920.8$ &  $720.4$ \\\midrule
哦\ 我\ 没有\ meeting\ 了 &  $92.2$ &  $75.9$ \\\midrule
okay kay\ 让\ 我\ 拿出\ 我\ 的\ calculator & $1260.3$ & $1284.6$ \\\midrule
the roomie lives in serangoon right & $2302.1$ & $1629.3$ \\\midrule
oh\ 他\ 拿\ third class\ 他\ 差\ 一点点\ 他的\ f. y. p. screwed up\ 他\ 拿\ 到\ b minus c plus & $299.7$ & $257.1$\\
\bottomrule
\end{tabular}}
\end{CJK*}









\vspace{5pt}
We observe that when less frequent words appear at switching points (like total, meeting, etc.), the DLM outperforms the mixed LM by a significant margin as illustrated in the first two sentences above. In cases of highly frequent words occurring at switching points, the DLM performs on par with or slightly worse than the mixed LM, as seen in the case of the third sentence. The DLM also performs slightly better within long stretches of monolingual text as seen in the fourth sentence. On the final sentence, which has multiple switches and long stretches of monolingual text, again the DLM performs better. As these examples illustrate, DLMs tend to show improved performance at less frequent switching points and within long stretches of monolingual text.

\paragraph{Effect of Trigrams.}
In standard monolingual datasets, trigram models consistently outperform bilingual models. However, in the SEAME corpus we did not find a pronounced difference between a bigram and a trigram model. This could be attributed to the fact that the number of highly frequent trigrams in our corpus is lower in comparison to that in the PTB dataset (Figure~\ref{fig:ngram}-b). 
As such, we have focused on bigram LMs in this work.

     


\section{Conclusions}

We introduced DLMs and showed robust improvements over mixed LMs in perplexity for code-switched speech. While the performance improvements for the ASR error rates are modest, they are achieved without the aid of any external language resources and without any computational overhead. We observe significant ASR improvements via lattice combination of DLMs and the standard mixed LMs. Future directions include investigating  properties of code-switched text which can be incorporated within DLMs, using monolingual data to enhance each DLM component and demonstrating the value of DLMs for multiple code-switched language pairs.

\paragraph{Acknowledgements.}
The last author gratefully acknowledges financial support from Microsoft Research India for this project, as well as access to Microsoft Azure cloud computing services.

\newpage 
\bibliographystyle{IEEEtran}
\bibliography{refs,mybib}

\end{document}